\documentclass{article}
\usepackage{ijcai23}

\usepackage{times}
\usepackage{soul}
\usepackage{url}
\usepackage[hidelinks]{hyperref}
\usepackage[utf8]{inputenc}
\usepackage[small]{caption}
\usepackage{graphicx}
\usepackage{amsmath}
\usepackage{amssymb}
\usepackage{amsthm}
\usepackage{booktabs}
\usepackage{algorithm}
\usepackage{algorithmic}
\usepackage[switch]{lineno}
\DeclareMathOperator*{\argmax}{arg\,max}
\DeclareMathOperator*{\onehot}{one\,hot}
\DeclareMathOperator*{\argmin}{arg\,min}

\title{MolHF: A Hierarchical Normalizing Flow for Molecular Graph Generation}

\author{
Yiheng Zhu$^1$\and
Zhenqiu Ouyang$^2$\and
Ben Liao$^3$\and
Jialu Wu$^4$\and
Yixuan Wu$^5$\and
Chang-Yu Hsieh$^{4}$\footnote{Corresponding authors.}\and
Tingjun Hou$^{4}$\footnotemark[1]\And
Jian Wu$^{5,6}$\footnotemark[1]\\
\affiliations
$^1$College of Computer Science and Technology, Zhejiang University\\
$^2$Polytechnic Institute, Zhejiang University\\
$^3$Tencent Quantum Laboratory, Tencent\\
$^4$College of Pharmaceutical Sciences, Zhejiang University\\
$^5$School of Public Health, Zhejiang University\\
$^6$Second Affiliated Hospital School of Medicine, Zhejiang University
\emails
\{zhuyiheng2020, oyzq, jialuwu, wyx\_chloe, kimhsieh, tingjunhou, wujian2000\}@zju.edu.cn,
liao@hotmail.co.uk
}

\begin{document}

\maketitle

\begin{abstract}
Molecular \textit{de novo} design is a critical yet challenging task in scientific fields, aiming to design novel molecular structures with desired property profiles. Significant progress has been made by resorting to generative models for graphs. However, limited attention is paid to hierarchical generative models, which can exploit the inherent hierarchical structure (with rich semantic information) of the molecular graphs and generate complex molecules of larger size that we shall demonstrate to be difficult for most existing models. The primary challenge to hierarchical generation is the non-differentiable issue caused by the generation of intermediate discrete coarsened graph structures. To sidestep this issue, we cast the tricky hierarchical generation problem over discrete spaces as the reverse process of hierarchical representation learning and propose MolHF, a new hierarchical flow-based model that generates molecular graphs in a coarse-to-fine manner. Specifically, MolHF first generates bonds through a multi-scale architecture, then generates atoms based on the coarsened graph structure at each scale. We demonstrate that MolHF achieves state-of-the-art performance in random generation and property optimization, implying its high capacity to model data distribution. Furthermore, MolHF is the first flow-based model that can be applied to model larger molecules (polymer) with more than 100 heavy atoms. The code and models are available at \url{https://github.com/violet-sto/MolHF}.
\end{abstract}

\section{Introduction}

Designing novel and valuable molecular structures is a critical yet challenging task with great application potential in drug and material discovery. Owing to the enormously vast chemical space with discrete molecular structures, whose scale was estimated to be on the order of $10^{33}$\cite{polishchuk2013estimation}, it is time-consuming and costly to search exhaustively in this space. Recently, there has been a surge of interest in developing generative models for graphs to resolve this issue by exploring the chemical space in a data-driven manner and sampling novel molecules~\cite{jin2018junction}.

Much prior work represents molecules as attributed graphs and leverages a variety of deep generative frameworks for molecular graph generation, including variational autoencoders (VAEs)~\cite{simonovsky2018graphvae}, generative adversarial networks (GANs)~\cite{de2018molgan}, and normalizing flows~\cite{shi2020graphaf,zang2020moflow}. However, most of them ignore the inherent hierarchical structure (with rich semantic information) of the molecular graphs. The significance of exploiting hierarchical structure lies in the fact that some particular substructures, such as molecular fragments and functional groups, appear frequently in molecules and are likely to determine molecular properties. Thereby, these models are sub-optimal for molecular generation and property optimization. Additionally, molecular graph generation is prone to errors, e.g., randomly removing or adding a covalent bond probably yields an invalid molecular structure. Thus, whilst existing models have demonstrated promising performance for small molecules, they tend to perform dreadfully for large molecules or even fail to effectively generate more complex molecules of larger size~\cite{jin2020hierarchical}. Recently, substructure-based methods~\cite{jin2020hierarchical}, where the substructures are manually extracted beforehand and leveraged as building blocks, have been proposed to exploit the implicit hierarchical structure. Nevertheless, the substructure extraction strategies are mostly based on heuristics and prior domain knowledge ~\cite{jin2018junction,jin2020hierarchical}, and the extraction process is cumbersome (e.g., counting the frequency of occurrence of all possible substructures). These practical drawbacks limit the flexibility and scalability of such methods.

\begin{figure*}[t]
\centering
\includegraphics[width=0.8\textwidth]{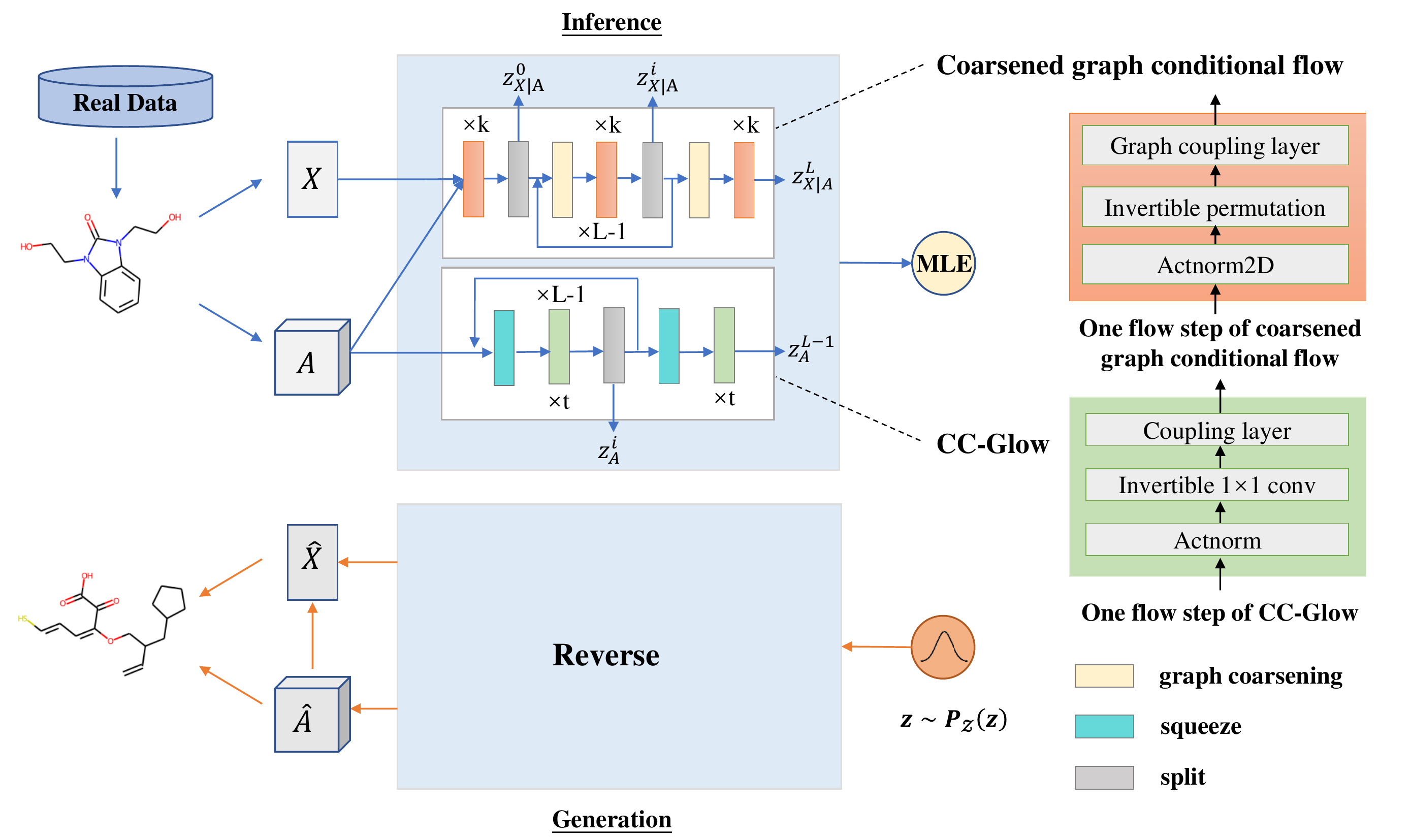} 
\caption{The framework of MolHF. A molecular graph $G$ is represented with atom matrix $X$ and bond tensor $A$. MolHF consists of two flows, namely a \textit{coarsened graph conditional flow} $f_{X \vert A}$ for hierarchically transforming $X$ given $A$, and a \textit{CC-Glow} $f_A$ for hierarchically transforming A. Right: the details of one flow step of $f_{X \vert A}$ and $f_A$.}
\label{fig:framework}
\end{figure*}

Advances in hierarchical generative frameworks have made significant progress in high-resolution image generation~\cite{karras2018progressive,yu2020wavelet} and super-resolution~\cite{liang2021hierarchical}, with the merits of better quality, scalability, and stability. Their key motivation is to progressively generate higher resolution images based on lower resolution input. However, applying hierarchical generative frameworks to derive molecular graphs is non-trivial. The primary challenge unique to hierarchical graph generation is generating intermediate discrete coarsened graph structures, which leads to the inability to backpropagate through samples. Currently, hierarchical molecular graph generation is still in its infancy, and it remains unclear how exploiting the hierarchical structure of molecular graphs could enhance the performance of generative models.

To overcome the above issues, we cast the tricky problem of hierarchical generation over discrete spaces as the reverse process of hierarchical representation learning resorting to invertible normalizing flows~\cite{dinh2014nice,dinh2016density}, and propose a new hierarchical flow-based generative model called MolHF, which generates molecular graphs in a coarse-to-fine manner. By defining an invertible transformation to capture the hierarchical structural information in the inference (encoding) process, MolHF can also exploit it in the reverse generation (decoding) process. Specifically, we propose \textit{CC-Glow}, a Criss-Cross network~\cite{huang2019ccnet} based multi-scale Glow~\cite{kingma2018glow} which offers a better relational inductive bias, to generate bonds at first, and a \textit{coarsened graph conditional flow} to generate atoms by performing graph convolution~\cite{kipf2017semi,schlichtkrull2018modeling} on the pre-generated coarsened graph structure (bonds) of each scale. We evaluate our MolHF in aspects of random generation and property optimization on the standard ZINC250K dataset~\cite{irwin2012zinc}. Further, we use the Polymer dataset~\cite{st2019message} to verify that MolHF is capable of modeling larger molecules that are impractical with prior flow-based methods. As demonstrated with extensive experiments, MolHF achieves state-of-the-art performance, indicating the effectiveness of the proposed hierarchical architecture. Finally, we analyze the generative mechanism of MolHF through hierarchical visualization. Our key contributions are summarized as follows.

\begin{itemize}
    \item We propose MolHF for hierarchical molecular graph generation. MolHF is the first flow-based model that can exploit the hierarchical structural information.
    \item We cast the non-trivial hierarchical generation problem over discrete spaces as the reverse process of hierarchical representation learning to sidestep the non-differentiable issue.
    \item MolHF outperforms state-of-the-art baselines in random generation and property optimization. Meanwhile, experiments on the Polymer dataset show that MolHF is the first flow-based model that can model larger molecules with more than 100 heavy atoms.
\end{itemize}

\section{Related Work}
\subsection{Normalizing Flows}
Normalizing flows are a class of probabilistic generative models that aim to learn invertible and differentiable mappings between complex target distributions and simple prior distributions. Compared with the prevailing generative models, such as VAEs and GANs, normalizing flows possess the merits of exact latent variable inference and likelihood estimation, efficient sampling, and stable training~\cite{kingma2018glow}. Advances in normalizing flows have made impressive progress in various applications, such as density estimation~\cite{dinh2016density}, image generation~\cite{kingma2018glow}, and variational inference~\cite{rezende2015variational}. Recently, normalizing flows also found their way into generating more valid and novel molecular graphs~\cite{shi2020graphaf}. Furthermore, much work has been devoted to improving normalizing flows, ranging from expressive power~\cite{ho2019flow++} to numerical stability~\cite{behrmann2021understanding}. For a more comprehensive review of normalizing flows, we refer the readers to ~\cite{papamakarios2021normalizing}.

\subsection{Molecular Graph Generation}
\label{sec:related}
Over the past few years, there has been a surge in generating molecular graphs resorting to various deep generative frameworks, e.g., VAEs~\cite{kingma2013auto}, GANs~\cite{goodfellow2014generative}, and autoregressive models. For example, JT-VAE~\cite{jin2018junction} first generates a junction tree that represents the scaffold over chemical substructures, then assembles them into a valid molecular graph. MolGAN~\cite{de2018molgan} is the first GAN-based model but shows poor performance in generating unique molecules. GCPN~\cite{you2018graph} formulates molecular graph generation as a sequential decision process generating the atoms and bonds alternately and leverages reinforcement learning to optimize the generated molecules.

There are also several flow-based models for molecular graph generation, which can be mainly grouped into two categories, i.e., autoregressive methods and one-shot methods. GraphAF~\cite{shi2020graphaf} and GraphDF~\cite{luo2021graphdf} adapt autoregressive flows~\cite{papamakarios2017masked} to generate atoms and bonds alternately in a sequential manner. GraphNVP~\cite{madhawa2019graphnvp} and MoFlow~\cite{zang2020moflow} are representative one-shot methods, which generate atoms conditioned on the pre-generated bonds in a two-step manner. Our MolHF follows the latter approach to exploit the coarsened graph structure for hierarchical generation while ensuring reversibility. The most related to our method is MolGrow~\cite{kuznetsov2021molgrow}, which starts with a single node and splits every node into two recursively to generate the entire molecule. Our MolHF differs from MolGrow in two main aspects: 1) MolGrow transforms the atom matrix and bond tensor simultaneously, therefore the transformed bond tensor becomes continuous hidden states which cannot provide structural information for modeling atoms in the coarsened scale. In contrast, MolHF is able to perform graph convolution to capture the hierarchical structural information, resorting to the aforementioned two-step approach. 2) MolGrow is less flexible and inefficient as the number of atoms in the training molecular graphs must be padded to be a power of 2, leading to extra computational costs. In contrast, we propose a graph coarsening operation to merge any number of nodes and compute the coarsened graph structure, making MolHF adaptable to various molecular datasets by setting appropriate coarsening ratios.

\section{Problem Statement and Background}
\subsection{Hierarchical Molecular Graph Generation}
The hierarchical molecular graph generation problem seeks to design a network (e.g., the decoder of VAE or generator of GAN) to generate the entire molecular graph in a coarse-to-fine manner. Put differently, the molecular graph is refined with increasingly fine details, conditioned on the coarser structure. Unfortunately, applying hierarchical generative frameworks to derive molecular graphs is non-trivial, since graphs are discrete objects. The primary challenge is generating intermediate discrete coarsened graph structures, which leads to the inability to backpropagate through samples. To sidestep this non-differentiable issue, we resort to normalizing flows~\cite{dinh2014nice} which construct a generative model by defining an invertible transformation. From the perspective of normalizing flows, the challenging problem of hierarchical molecular graph generation can be regarded as the reverse process of hierarchical representation learning.

For a molecular graph $G = (X,A)$ with $n$ atoms, we represent it with an atom matrix $X \in \{0,1\}^{n \times d}$ and a bond tensor $A \in \{0,1\}^{b \times n \times n}$, where $d$ and $b$ are the number of different types of atoms and bonds, respectively. The element $X_{i,:}$ and $A_{:,i,j}$ are one-hot vectors indicating the type of the $i^{th}$ atom and the type of the bond between the $i^{th}$ atom and $j^{th}$ atom, respectively. Our ultimate purpose is to learn the distribution $P(G)$ via a hierarchical probabilistic generative model.

\subsection{Normalizing Flows}
The flow-based generative models define a parameterized invertible transformation $f_\theta: \mathcal{X} \rightarrow \mathcal{Z}$ from observational data $x \sim P_\mathcal{X}(x)$ to latent variables $z \sim P_\mathcal{Z}(z)$, where $P_\mathcal{X}(x)$ is a complex target distribution and $P_\mathcal{Z}(z)$ is a simple base distribution (e.g., standard isotropic Gaussian distribution). Under the change-of-variables formula, the exact log-likelihood of $x$ is tractable:

\begin{equation}
    \log P_\mathcal{X}(x)= \log P_\mathcal{Z}(f_\theta(x)) + \log \left|\text{det}\frac{\partial f_\theta(x)}{\partial x}\right|
\end{equation}
where $\text{det}\frac{\partial f_\theta(x)}{\partial x}$ is the Jacobian determinant of $f_\theta$ at $x$. Thereby, unlike VAEs, the flow-based models can be trained directly with the maximum likelihood estimation. As for generation, a sample is efficiently generated via the inverse transformation $x=f^{-1}_\theta(z)$ where $z$ is sampled from $P_{\mathcal{Z}}(z)$.

In practice, $f_\theta = f_1 \circ \cdots \circ f_L$ is composed of a sequence of invertible functions whose Jacobian determinant is convenient to calculate. Herein, we consider using the affine coupling layer, actnorm, and invertible permutation.

\paragraph{Affine coupling layer.} One of the most widely used invertible functions is the affine coupling layer~\cite{dinh2016density}. Formally, given $x \in \mathbb{R}^D$ and $d < D$, the output $y \in \mathbb{R}^D$ can be computed as follows:
\begin{align}
    y_{1:d}& = x_{1:d} \\
    y_{d+1:D}& = x_{d+1:D} \odot r(s(x_{1:d}))+t(x_{1:d})
\end{align}
where $\odot$ stands for Hadamard product, $s$ and $t$ are scale and translation functions, respectively, and $r$ is a rescale function that is commonly instantiated as an exponential function. Since the log-determinant of the triangular Jacobian of this transformation is efficiently computed as $\sum_j \log r(s(x_{1:d}))_j$, $s$ and $t$ can be arbitrarily complex functions to improve the expressive power~\cite{dinh2016density}. To mitigate numerical instability, we adopt the sigmoid function following the Swish activation function~\cite{ramachandran2017searching} as $r$ with a restricted range, whose effect was thoroughly investigated in ~\cite{behrmann2021understanding}.

\paragraph{Actnorm.} Actnorm~\cite{kingma2018glow} is proposed to not only alleviate the problems associated with the numerical instability but also sidestep the noise introduced by the batch normalization. The main idea of actnorm is to normalize the activations to zero mean and unit variance before starting training. Specifically, the parameters of actnorm are initialized with the per-channel mean and standard deviation of the first batch of data, and optimized as regular parameters. As for the feature matrix, actnorm is extended to normalize the feature dimension. For consistency, we continue to call this variant actnorm2D~\cite{zang2020moflow}.

\paragraph{Invertible permutation.} Invertible $1 \times 1$ convolution with weight matrix $W \in \mathbb{R}^{c \times c}$ is a learnable permutation proposed to replace the fixed random permutation~\cite{kingma2018glow}. With $1 \times 1$ convolution, all channels are allowed to influence one another, and the expressive power of the following affine coupling layers is subsequently improved. Different from 3D tensor (images), the feature matrix can be permuted by directly multiplying $W$. To reduce the cost of computing log-determinant of $W$, we formulate it with LU decomposition~\cite{kingma2018glow}.

\section{Methods}
In this section, we first introduce the overview of our proposed MolHF, then we elucidate the specifics of MolHF. Lastly, we describe the generation process of MolHF.

\subsection{Overview of MolHF}
Similar to existing one-shot flow-based models such as GraphNVP~\cite{madhawa2019graphnvp} and MoFlow~\cite{zang2020moflow}, MolHF aims to generate molecular graphs in one pass through the model, but in a coarse-to-fine manner. To exploit the coarsened graph structure for hierarchical generation while ensuring the reversibility of MolHF, we conditionally factorize the distribution $P(G)$ according to the combinatorial structure of molecular graphs, giving
\begin{equation}
    P(G) = P(X,A) = P(A)P(X|A)
\end{equation}
Based on this factorization, MolHF wishes to hierarchically transform the atom matrix $X$ and bond tensor $A$ respectively. As illustrated in Figure~\ref{fig:framework}, MolHF consists of two flows, namely a \textit{coarsened graph conditional flow} $f_{X \vert A}$ for atoms and a \textit{CC-Glow} $f_A$ for bonds. We adopt the well-established dequantization technique~\cite{shi2020graphaf,zang2020moflow} to convert both $X$ and $A$ into continuous data by adding uniform noise, since discrete data do not fit into the change-of-variables formula. Under the maximum likelihood principle, the parameters of MolHF are updated by gradient descent to minimize the negative log-likelihoods $ \log P(G)$ over all training molecular graphs:

\begin{equation}
     \resizebox{.91\linewidth}{!}{$
            \displaystyle\argmin_{f_A,f_{X \vert A}} \mathbb{E}_{X,A\sim P_{data}} [-\log P(A;f_A) - \log P(X \vert A;f_{X \vert A})]
            $}
\end{equation}


\subsection{Coarsened Graph Conditional Flow}
Conditioning on the bond tensor $A$, \textit{coarsened graph conditional flow} $f_{X \vert A}$ aims to learn $P(X \vert A)$ by transforming the atom matrix $X$ into latent variables $z_{X \vert A}=f_{X \vert A}(X,A)$ in a hierarchical manner. Inspired by this motivation, we propose a graph coarsening operation and corresponding coarsened graph coupling layers to extend the multi-scale architecture proposed in ~\cite{dinh2016density}.

\paragraph{Multi-scale architecture.} At each scale, we perform three successive operations. Firstly, a graph coarsening operation is applied to a fine-grained molecular graph, except at the finest scale. Secondly, several steps of transformation which contains actnorm2D followed by an invertible permutation, followed by a coarsened graph coupling layer, are utilized to extract the coarse-grained features. Thirdly, half the features are split into latent variables, except at the coarsest scale. This procedure can be defined recursively,

\begin{align}
    h^{1},z^{0}_{X \vert A},A^{0} &= f_{X \vert A}^{1}(h^{0},A^{0}) \\
    h^{i+1},z^{i}_{X \vert A},A^{i} &= f_{X \vert A}^{i+1}(h^{i},A^{i-1}) \\
    z^{L}_{X \vert A},A^{L} &= f_{X \vert A}^{L+1}(h^{L},A^{L-1}) \\
    z_{X \vert A} &= (z^{0}_{X \vert A},\dots,z^{L}_{X \vert A})
\end{align}
where $h^0=X$ and $A^0=A$. Note that the correlations between latent variables of different scales are modeled by assuming $z^i_{X \vert A} \sim \mathcal{N}(\mu_i(h^{i+1}),\sum_i(h^{i+1}))$. 

\paragraph{Graph coarsening operation.} We merge every $k_{i-1}$ nodes to construct the coarsened graph, where $k_{i-1}$ is a hyper-parameter controlling the degree of coarsening at each scale. The merged node features are obtained with concatenation, while the coarsened graph structure $A^i$ is computed as:

\begin{equation}
    A^{i}_{m,:,:} = S^TA^{i-1}_{m,:,:}S,\quad m = 1,\dots,b
\end{equation}
where $S \in \{0,1\}^{n_{i-1} \times \frac{n_{i-1}}{k_{i-1}}}$ is a predefined hard assignment matrix assigning every $k_{i-1}$ nodes as a cluster in order, $n_{i-1}$ is the number of nodes in the coarsened graph at the ${i-1}^{th}$ scale, and $A^{i}$ indicates the connectivity between each pair of clusters~\cite{ying2018hierarchical}. As a notorious problem in graph generation, it is tough to define the most suitable atom ordering. We use breadth-first search ordering to preserve spatial locality, which is vital for hierarchical modeling.

\paragraph{Coarsened graph coupling layers.} We transform the atom matrix via the affine coupling layers. To achieve permutation equivariance at each scale, we split atom features into two parts along the feature (column) dimension and adopt R-GCNs~\cite{schlichtkrull2018modeling} in the coupling networks, leveraging the coarsened graph structure. The convolution operation of R-GCN can be written as follows: 
\begin{equation}
  H^l = \sum_{i=1}^{b} D^{-1}A_iH^{l-1}W_i + H^{l-1}W_0
\end{equation}
where $H^l$ is the node embeddings in $l^{th}$ layer, $A_i$ is the $i^{th}$ slice of adjacency tensor, $D_{jj} = \sum_{b,i} A_{b,i,j}$, and $\{W\}_{i=0}^b$ are weight matrices. Compared with the graph coupling layers used in MoFlow~\cite{zang2020moflow} that split atom features along the row dimension, ours are more flexible since the number of coupling layers is decoupled from the number of nodes.

\subsection{CC-Glow}
\textit{CC-Glow} $f_A$ aims to learn $P(A)$ by transforming bond tensor $A$ into latent variables $z_A=f_A(A)$ in analogy with multi-channel images. We adopt the architecture of multi-scale Glow but propose a novel Criss-Cross Attention (CCA)~\cite{huang2019ccnet} based coupling layer to offer a better relational inductive bias for modeling bonds. 

\paragraph{CCA-based coupling layers.} 
For modeling bonds, it is essential to utilize the local information of the incident atoms. However, the atom matrix is unavailable at this stage in ensuring the reversibility of MolHF. Therefore, we decide to assign atoms with features aggregated from relevant bonds (attached to the atom) through message passing~\cite{gilmer2017neural}. Based on the above analysis, we achieve this goal by introducing CCA to extract features from the bond tensor in horizontal and vertical directions as the contextual messages of two atoms connected by a bond, respectively. See Appendix A.1 for more implementation details. We build the coupling network with three layers, where the first and last layers are $3 \times 3$ convolutions, while the center layer is CCA.

\begin{table*}
    \centering
    \scalebox{0.9}{\begin{tabular*}{\textwidth}{@{\extracolsep{\fill}}lrrrrr}
    \toprule
    Method  & Validity & Validity w/o correction & Uniqueness & Novelty & Reconstruct \\
    \midrule
    JT-VAE         & 100$\%$           & n/a              & 100$\%$              & 100$\%$            & 76.7$\%$ \\
    GCPN           & 100$\%$             & 20$\%$              & 99.97$\%$           & 99.97$\%$          & n/a  \\
    GraphAF        & 100$\%$           & 68$\%$               & 99.10$\%$            & 100$\%$             & 100$\%$  \\
    GraphDF        & 100$\%$           & 89.03$\%$            & 99.16$\%$            & 100$\%$              & 100$\%$
     \\
    MoFlow         & 100$\%$           & 81.76 $\pm$ 0.21$\%$ & 99.99$\%$ & 100$\%$              & 100$\%$  \\
    GraphCNF       & 96.35$\%$           & 83.41$\%$            & 99.99$\%$            & 100$\%$              & 100$\%$  \\
    MolGrow        & 100$\%$           & 57.80 $\pm$ 7.75$\%$ & 99.06$\%$ & 99.96$\%$ & 100$\%$  \\
    \midrule
    MolHF          & \textbf{100$\%$}           & \textbf{94.89 $\pm$ 0.20$\%$} & \textbf{100$\%$}              & \textbf{100$\%$}             & \textbf{100$\%$}  \\
    \bottomrule
    \end{tabular*}}
    \caption{Random generation and reconstruction performance on the ZINC250k dataset.}
    \label{tab:performance_zinc}
\end{table*}

\begin{table*}
    \centering
    \scalebox{0.9}{\begin{tabular*}{\textwidth}{@{\extracolsep{\fill}}lrrrrrr}
    \toprule
    Method  & Validity & Validity w/o correction & Validity w/ filter & Uniqueness & Novelty & Reconstruct \\
    \midrule
    JT-VAE      & 100$\%$ & n/a                            & n/a                         & 94.10$\%$ & - & 58.5$\%$ \\
    HierVAE     & 100$\%$ & n/a                            & n/a                         & 97.00$\%$ & - & 79.9$\%$ \\
    GraphAF     & 100$\%$ & 28.58 $\pm$ 2.98$\%$           &  1.74 $\pm$ 0.32$\%$       & 100$\%$  & 100$\%$ & 100$\%$  \\
    GraphDF     & 100$\%$ & 17.23 $\pm$ 0.11$\%$           &  0$\%$       & 98.89$\%$ & 100$\%$ & 100$\%$ \\
    MoFlow      & 100$\%$ & 0$\%$                        &  0$\%$       & 100$\%$ & 100$\%$ & 100$\%$ \\
    GraphCNF    & 25.32 $\pm$ 1.10$\%$                     &  19.32 $\pm$ 0.96$\%$  & 11.82 $\pm$ 1.65$\%$             & 100$\%$ & 100$\%$ & 100$\%$ \\
    \midrule
    MolHF       & \textbf{100$\%$} & \textbf{42.75 $\pm$ 0.47$\%$}          & \textbf{35.62 $\pm$ 0.43$\%$} & \textbf{100$\%$} & \textbf{100$\%$} & \textbf{100$\%$} \\   
    \bottomrule
    \end{tabular*}}
    \caption{Random generation and reconstruction performance on the Polymer dataset. Results of flow-based baselines are obtained by running their official source code. Note that we omit the results of MolGrow since the source code is not released.}
    \label{tab:performance_polymer}
\end{table*}

\subsection{Two-step Molecular Graph Generation}
Benefiting from the reversibility of MolHF, molecular graph generation is simply executed by drawing $z=(z_{X \vert A},z_A)$ from $P_{\mathcal{Z}}(z)$ and calling the inverse transformation of MolHF. In our two-step generative manner, the bond tensor $\hat{A}$ is generated by the inverse transformation of $f_A$ at first, then the atom matrix $\hat{X}$ is generated by the inverse transformation of $f_{X \vert A}$, conditioned on the pre-generated $\hat{A}$. Formally, this process can be summarized as,
\begin{equation}
    \label{eq:generation}
    \begin{aligned}
    \hat{A} &= \onehot(\argmax(\widetilde{A})), \quad \widetilde{A} = f_{A}^{-1}(z_A),  \\
    \hat{X} &= \onehot(\argmax(\widetilde{X})), \quad \widetilde{X} = f_{X\vert A}^{-1}(z_X, \hat{A})
    \end{aligned}
\end{equation}
where the composite operation $\onehot \cdot \argmax$ maps the dequantized continuous data back to the discrete one-hot data. Lastly, a post-hoc validity correction mechanism~\cite{zang2020moflow} is incorporated to ensure that the generated molecular graphs are valid. This generation process can exploit the hierarchical structural information captured by the inverse transformation of MolHF.

\section{Experiments}
\label{sec:exp}
Following previous works~\cite{jin2018junction,zang2020moflow}, we demonstrate the effectiveness of our MolHF in various aspects: generation and reconstruction, property optimization, and hierarchical visualization. 

\paragraph{Datasets.} We evaluate our method on the ZINC250K dataset~\cite{irwin2012zinc} for a fair comparison. In addition, we also use the Polymer dataset~\cite{st2019message} to verify that our method is capable of scaling to larger molecules. The ZINC250K dataset consists of 249,455 molecules with up to 38 atoms of 9 different types, and the Polymer dataset consists of 86,973 larger polymer-like molecules with up to 122 atoms of 7 different types. For the preprocessing strategy, we mostly follow ~\cite{zang2020moflow}, the difference is that we pad all the molecules in the ZINC250K and Polymer with 2 and 6 extra virtual atoms respectively, to ensure that each molecular graph can be coarsened over multiple scales. 

\paragraph{Baselines.} We compare MolHF with the following baselines. JT-VAE~\cite{jin2018junction} and HierVAE~\cite{jin2020hierarchical} are VAE-based models. GCPN~\cite{you2018graph} is an autoregressive model. GraphAF~\cite{shi2020graphaf} and GraphDF~\cite{luo2021graphdf} are autoregressive flow-based models. MoFlow~\cite{zang2020moflow}, GraphCNF~\cite{lippe2021categorical}, and MolGrow~\cite{kuznetsov2021molgrow} are one-shot flow-based models.

\paragraph{Implementation.} For the generation and reconstruction, the latent variables are assumed to follow a prior isotropic Gaussian distribution $\mathcal{N}(0,\sigma^2\text{I})$, where $\sigma$ is a learnable parameter denoting the standard deviation. For property optimization and hierarchical visualization, the same model trained on the ZINC250K dataset is used for all experiments. Further implementation details can be found in Appendix C.

\subsection{Generation and Reconstruction}
\paragraph{Setup.} We evaluate the ability of MolHF to generate and reconstruct molecular graphs via the widely used metrics: \textbf{Validity} and \textbf{Validity w/o correction} are the percentages of chemically valid molecules among all the generated graphs with and without validity correction, respectively. \textbf{Uniqueness} is the percentage of unique molecules among all the generated valid molecules. \textbf{Novelty} is the percentage of generated valid molecules that do not appear in the training set. \textbf{Reconstruction} is the percentage of molecules in the training set that can be reconstructed from latent variables. As for the experiments on the Polymer dataset, to verify that MolHF does generate more large molecules, we introduce \textbf{Validity w/ filter} to calculate the percentage of valid large molecules with more than 38 atoms (the maximum number of atoms in the ZINC250K dataset) among all the generated graphs without validity correction. In practice, higher-quality samples are drawn from the reduced-temperature models~\cite{kingma2018glow}. To control the trade-off between quality and diversity, we sample from the distribution $\mathcal{N}(0,(t\sigma)^2\text{I})$ with the temperature $t<1$, similar to MoFlow and MolGrow. To be comparable with the aforementioned baselines, we sample 10,000 molecular graphs and report the mean and standard deviation of results over 5 runs.


\paragraph{Results.} Table~\ref{tab:performance_zinc} shows that MolHF outperforms the state-of-the-art baselines on all the metrics for the ZINC250K dataset. Thanks to the validity correction mechanism, MolHF achieves $100\%$ \textbf{Validity}. We argue that \textbf{Validity w/o correction} can better assess how well models capture the data distribution, since the corrected graphs can be completely different from the original model outputs. Compared with one-shot flow-based models MoFlow and GraphCNF, MolHF attains an increase of $13.1\%$ and $11.5\%$ in \textbf{Validity w/o correction} respectively, implying the superiority of modeling the molecular graphs with hierarchical architecture. Compared with another hierarchical flow-based model MolGrow, MolHF outperforms it by a large margin ($37.1\%$) with respect to \textbf{Validity w/o correction}, indicating that exploiting the coarsened graph structure is crucial for enhancing the performance of hierarchical generative models. Owing to the invertible nature, MolHF is able to reconstruct all molecules, whereas JT-VAE and GCPN can not. Furthermore, experiments on the Polymer dataset demonstrate that MolHF can be applied to model larger molecules that are impractical with previous flow-based models, as shown in Table~\ref{tab:performance_polymer}. We find that the performance of all flow-based baselines degrades substantially when modeling larger molecules. MoFlow even cannot fit the data distribution to generate valid molecules. While GraphAF and GraphDF can generate a few small molecules in the early stage of the autoregressive generation process, they become fallible as the current molecules get larger so that only a tiny fraction of the valid molecules are large. GraphCNF, the best flow-based baseline, only generates 11.8$\%$ large molecules, which is still unsatisfactory. By contrast, MolHF can generate approximately 43$\%$ valid molecules, most of which are large. These results imply that the hierarchical architecture exploiting the coarsened graph structures enables MolHF to model the more complex distribution by better capturing local and global information of large molecules. Compared with the substructure-based model HierVAE, MolHF generates more unique molecules and reconstructs all molecules. We illustrate several valid large molecules (with rich topological complexity) generated by MolHF in Appendix B.1.

\paragraph{Empirical running time.}
MolHF is significantly more efficient in aspects of training and sampling. For the ZINC250K dataset, the reported running time of MoFlow and MolGrow is 22 hours and 2 days respectively, while MolHF converges faster and can get the results within 12 hours. For the Polymer dataset, due to the huge memory cost for parallel training, autoregressive GraphAF and GraphDF cost roughly 50 minutes/epoch to train, while MolHF is $3 \times$ faster. Besides, GraphAF and GraphDF cost 2 hours and 20 minutes to randomly sample 10,000 molecules (without validity correction) respectively, whereas MolHF only takes 50 seconds, $144 \times$ and $24 \times$ faster. The efficiency is compared on the same computing facilities using 1 Tesla V100 GPU.

\begin{table}
\centering
\begin{tabular}{lllll}
\toprule
Method  & 1st & 2nd & 3rd & 4th \\
\midrule
ZINC250K (dataset)    & 0.948  & 0.948  & 0.948 & 0.948 \\
\midrule
JT-VAE  & \textbf{0.948}  & 0.911  & 0.911 & - \\
GCPN    & \textbf{0.948}  & 0.947  & 0.946 & -   \\
GraphAF & \textbf{0.948}  & \textbf{0.948}  & 0.947 & 0.946  \\
GraphDF & \textbf{0.948}  & \textbf{0.948}  & \textbf{0.948} & -  \\
MoFlow  & \textbf{0.948}  & \textbf{0.948}  & \textbf{0.948} & \textbf{0.948}   \\
MolGrow & \textbf{0.948}  & \textbf{0.948}  & \textbf{0.948} & -   \\
\midrule
MolHF   & \textbf{0.948}  & \textbf{0.948}  & \textbf{0.948} & \textbf{0.948}   \\
\bottomrule
\end{tabular}
\caption{Discovered novel molecules with the best QED scores.}
\label{tab:unconstrained}
\end{table}

\begin{table}
\centering
\scalebox{0.6}{
\begin{tabular}{lllrlllr}
\toprule
    & \multicolumn{3}{c}{JT-VAE}  
    & \multicolumn{3}{c}{GCPN}  \\
\cline{2-4}\cline{6-8}
$\delta$   & \textbf{Improvement} & \textbf{Similarity} & \textbf{Success} & & \textbf{Improvement} & \textbf{Similarity} & \textbf{Success} \\
\hline
\textbf{0.0} & 1.91$\pm$2.04 & 0.28$\pm$0.15 & 97.5$\%$ & & 4.20$\pm$1.28    & 0.32$\pm$0.12 & 100.0\%   \\
\textbf{0.2} & 1.68$\pm$1.85 & 0.33$\pm$0.13 & 97.1$\%$ & & 4.12$\pm$1.19    & 0.34$\pm$0.11 & 100.0\%    \\
\textbf{0.4} & 0.84$\pm$1.45 & 0.51$\pm$0.10 & 83.6$\%$ & & 2.49$\pm$1.30    & 0.48$\pm$0.08 & 100.0\%    \\
\textbf{0.6} & 0.21$\pm$0.71 & 0.69$\pm$0.06 & 46.4\% & & 0.79$\pm$0.63    & 0.68$\pm$0.08 & 100.0\%    \\
\hline
    & \multicolumn{3}{c}{GraphDF}         & \multicolumn{3}{c}{MolHF}          \\
\cline{2-4}\cline{6-8}
$\delta$    & \textbf{Improvement} & \textbf{Similarity} & \textbf{Success} & & \textbf{Improvement} & \textbf{Similarity} & \textbf{Success} \\
\hline
\textbf{0.0}  & 5.93$\pm$1.97 & 0.30$\pm$0.12  & 100.0$\%$ & & \textbf{6.98$\pm$3.37} & 0.25$\pm$0.13 & 100.0$\%$ \\
\textbf{0.2}  & 5.62$\pm$1.65 & 0.32$\pm$0.10  & 100.0$\%$ & & 
\textbf{6.29$\pm$3.36} & 0.32$\pm$0.13  & 99.8$\%$  \\
\textbf{0.4}  & \textbf{4.13$\pm$1.41}  & 0.47$\pm$0.07  & 100.0$\%$ & & 3.97$\pm$2.78  & 0.52$\pm$0.12  & 90.9$\%$  \\
\textbf{0.6}  & 1.72$\pm$1.15  & 0.67$\pm$0.06  & 93.0$\%$ & & \textbf{2.55$\pm$2.88}  & 0.72$\pm$0.11  & 55.9$\%$  \\
\bottomrule
\bottomrule
\end{tabular}}
\caption{Constrained optimization performance on the test set.}
\label{tab:constrained_test}
\end{table}

\begin{table}
\centering
\scalebox{0.6}{
\begin{tabular}{lllrlllr}
\toprule
    & \multicolumn{3}{c}{MoFlow} 
    & & \multicolumn{3}{c}{MolGrow}\\
\cline{2-4}\cline{6-8}
$\delta$   & \textbf{Improvement} & \textbf{Similarity} & \textbf{Success} & & \textbf{Improvement} & \textbf{Similarity} & \textbf{Success} \\
\hline
\textbf{0.0} & 8.61$\pm$5.44 & 0.30$\pm$0.20 & 98.88$\%$ & & 14.84$\pm$5.79 & 0.05$\pm$0.04 & 100.00$\%$ \\
\textbf{0.2} & 7.06$\pm$5.04  & 0.43$\pm$0.20  & 96.75$\%$ & & \textbf{11.99$\pm$6.45} & 0.23$\pm$0.05  & 99.80$\%$  \\
\textbf{0.4} & 4.71$\pm$4.55  & 0.61$\pm$0.18  & 85.75$\%$ & & \textbf{8.34$\pm$6.85}  & 0.44$\pm$0.05  & 99.80$\%$ \\
\textbf{0.6} & 2.10$\pm$2.86   & 0.79$\pm$0.14  & 58.25$\%$ & & 4.01$\pm$5.61  & 0.65$\pm$0.07  & 97.70$\%$  \\
\hline
    & \multicolumn{3}{c}{GraphAF} 
    & & \multicolumn{3}{c}{MolHF}\\
\cline{2-4}\cline{6-8}
$\delta$   & \textbf{Improvement} & \textbf{Similarity} & \textbf{Success} & & \textbf{Improvement} & \textbf{Similarity} & \textbf{Success} \\
\hline
\textbf{0.0} & 13.13$\pm$6.89 & 0.29$\pm$0.15  & 100.00$\%$ & & \textbf{15.22$\pm$5.99} & 0.15$\pm$0.16 & 100.00$\%$   \\
\textbf{0.2} & 11.90$\pm$6.86 & 0.33$\pm$0.12  & 100.00$\%$ & & 11.02$\pm$6.20 & 0.36$\pm$0.20  & 98.38$\%$  \\
\textbf{0.4} & 8.21$\pm$6.51  & 0.49$\pm$0.09  & 99.88$\%$ & & 7.18$\pm$7.06  & 0.61$\pm$0.18  & 86.13$\%$ \\
\textbf{0.6} & 4.98$\pm$6.49  & 0.66$\pm$0.05  & 96.88$\%$ & & \textbf{6.46$\pm$7.67}  & 0.78$\pm$0.13  & 63.25$\%$ \\
\bottomrule
\bottomrule
\end{tabular}}
\caption{Constrained optimization performance on the entire set.}
\label{tab:constrained_entire}
\end{table}

\begin{figure}[t]
    \centering
    \includegraphics[width=0.45\textwidth]{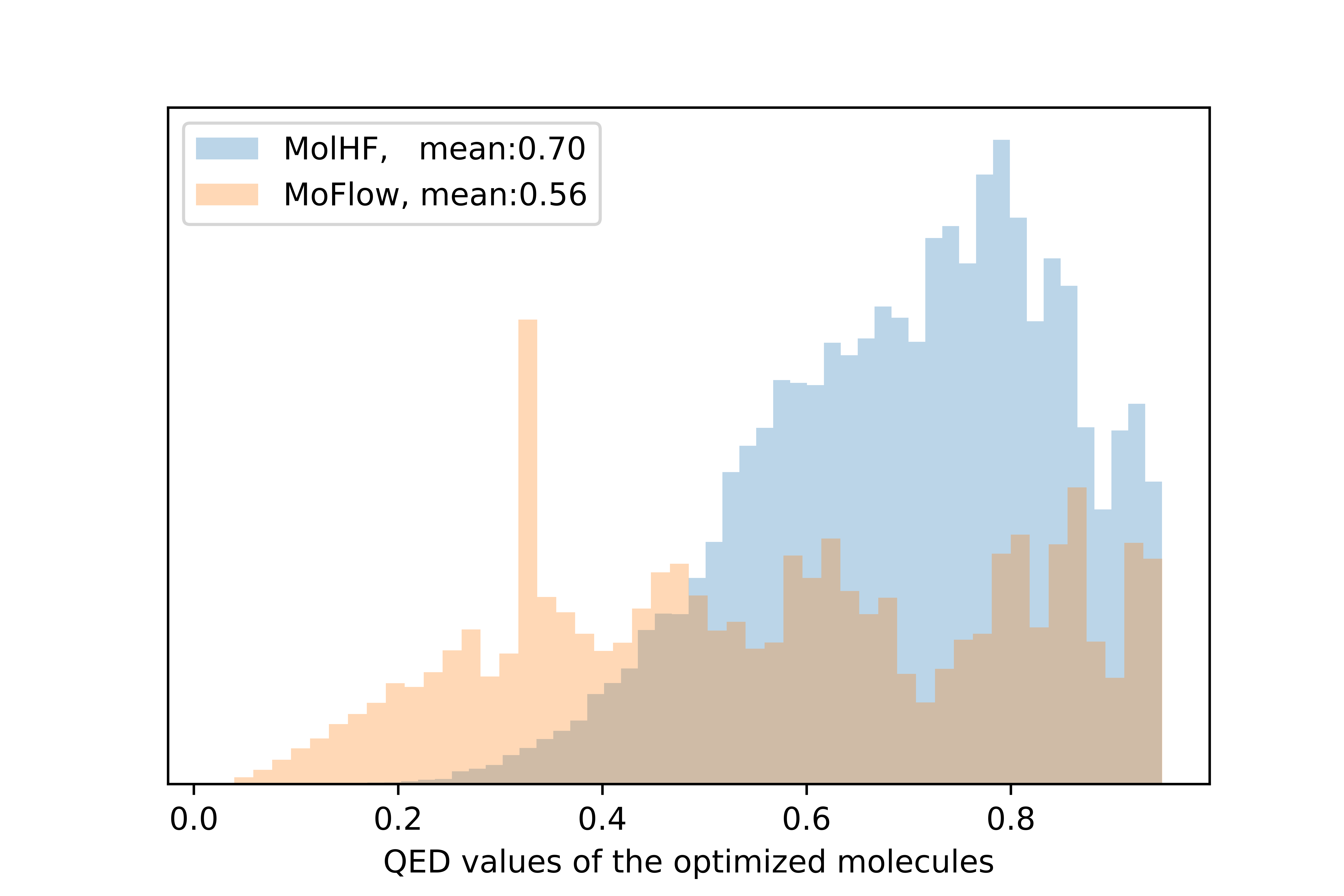}
    \caption{The histogram of the optimized QED.}
    \label{fig:hist_QED}
\end{figure}

\subsection{Property Optimization}
\paragraph{Setup.} The unconstrained property optimization task aims at discovering novel molecules with the best \textbf{QED}, which measures the druglikeness of the molecules. The constrained property optimization task aims at modifying the given molecules to improve \textbf{PlogP}, which denotes logP score penalized by ring size and synthetic accessibility, with the constraint that the Tanimoto similarity of Morgan fingerprint between the modified and original molecules is above a threshold $\delta$~\cite{jin2018junction}. We formulate property optimization as a latent space optimization (LSO) problem~\cite{gomez2018automatic}. Formally, we first train a surrogate network $h$ which takes as input the latent variables $z$ encoded with the trained MolHF to predict the property values, then perform gradient ascent $z \gets z + \alpha \cdot \nabla_z h$ iteratively to navigate the latent space. We decode the latest latent variables with MolHF to obtain the optimized molecule. For constrained property optimization, we start from the latent variables of 800 molecules with the lowest \textbf{PlogP}~\cite{jin2018junction}. Note that two different sets of molecules selected from the test set~\cite{jin2018junction} or the entire set~\cite{shi2020graphaf} are used in prior methods. We thus report results separately.

\paragraph{Results.} We show the best scores discovered in the unconstrained optimization task in Table~\ref{tab:unconstrained}. Like MoFlow, MolHF finds more novel molecules with the best QED score (0.948) appearing in the dataset. Further, we illustrate the histogram of the optimized property scores in Figure~\ref{fig:hist_QED} to verify that MolHF finds molecules with higher QED scores than MoFlow on average (0.70 vs.~0.56). As shown in Table~\ref{tab:constrained_test} and Table~\ref{tab:constrained_entire}, MolHF achieves higher property improvement than GraphDF under 3 of 4 similarity constraints and outperforms GraphAF and MolGrow under 2 of 4 similarity constraints, while maintaining higher similarities. Compared with the autoregressive GCPN, GraphDF, and GraphAF which fine-tune the generator with reinforcement learning and optimize each molecule hundreds of times, the optimization strategy of MolHF is more conveniently implementable and efficient. This nonetheless results in the issues of success rate and robustness. Fortunately, these issues can be alleviated with the recent advanced LSO approaches~\cite{tripp2020sample,notin2021improving}, potentially leading to better performance.
\label{sec:exp_optimization}

\subsection{Hierarchical Visualization}
\label{sec:visualization}
To investigate the generative mechanism of MolHF, we visualize the generated molecules in terms of the semantics of the learned latent variables and the generated substructures.

\begin{figure}[t]
    \centering
    \includegraphics[width=0.45\textwidth]{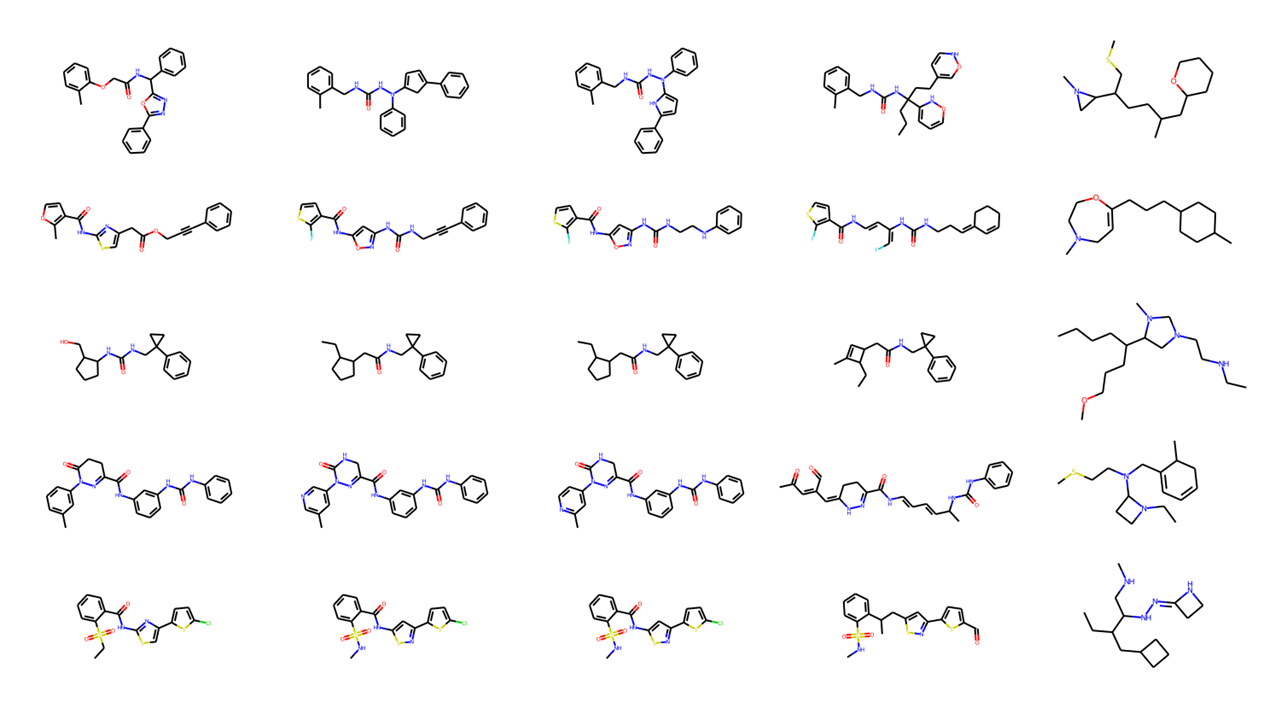}
    \caption{The leftmost column represents the original molecules, the subsequent columns are obtained by resampling the latent variables of coarser scale, resampling more and more as we go right.}
    \label{fig:resample_zinc250k}
\end{figure}

\paragraph*{Semantics of the learned latent variables.}
We infer the latent variables and resample the latent variables of the finest scale, increasing the coarsest scale affected by this resampling~\cite{dinh2016density}. We decode the resampled latent variables and illustrate several cases in Figure~\ref{fig:resample_zinc250k}, where the leftmost column represents the original molecules, and the subsequent columns (from left to right) are obtained by resampling the latent variables of coarser scale, resampling more and more as we go right. As we can see, latent variables of finer scale affect the local structure (atom and bond), while those of coarser scale affect the global structure (skeletal structure). 

\begin{figure}
\centering    
\includegraphics[width=0.45\textwidth]{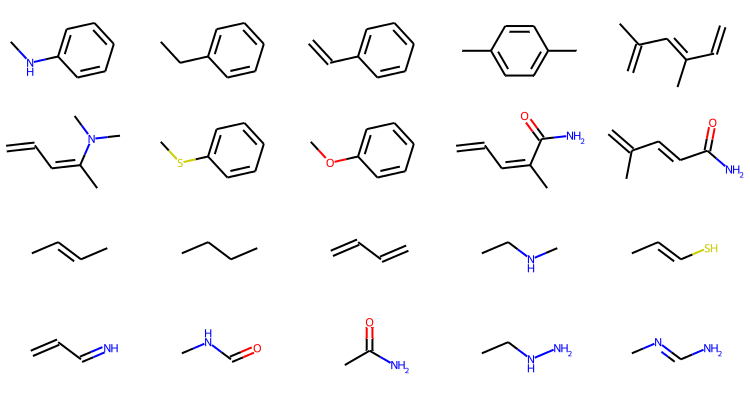} 
\caption{Illustration of the most frequent substructures generated by MolHF.}
\label{fig:substructures}     
\end{figure}

\paragraph*{Generated substructures.}
In Figure~\ref{fig:substructures}, we illustrate some of the most frequent substructures generated by MolHF. These learned substructures are chemically meaningful, demonstrating that MolHF does capture the rich valid structure and semantic information of molecular graphs. By comparison, we find that these substructures (such as 2-butylene, methylethylamine, and acetamide, to name a few) often appear in the fragment vocabulary~\cite{jin2018junction,jin2020hierarchical,xie2021mars,kong2022molecule} manually extracted based on frequency of occurrence. Thus, we argue that MolHF successfully extracts common substructures from the data without any need for domain knowledge, and matches the prior distribution of fragments to some extent. Meanwhile, due to the hierarchical generation, some of the substructures generated by MolHF are composed of smaller structural motifs. For instance, the substituted benzene rings can be further divided according to the retrosynthesis-based BRICS algorithm~\cite{degen2008art}. To link coarsened graph structure with substructures, we decompose several molecules with the same coarsened graph structure in Appendix B.2, where atoms are colored according to the substructures to which they belong. We observe that MolHF learns to hierarchically generate molecular graphs in a coarse-to-fine manner by first generating a coarsened graph structure (skeletal structure) and then refining the substructures.

\section{Discussion and Conclusion}
\label{sec:discussion}
In this paper, we present MolHF, a hierarchical flow-based model to generate molecular graphs in a coarse-to-fine manner. As demonstrated with extensive experiments, MolHF achieves state-of-the-art performance in random generation and property optimization, while being capable of modeling larger molecules that are impractical with prior flow-based methods. We hope that our MolHF can spur further research to unleash the great potential of hierarchical generative models in the graph generation field.

\paragraph{Limitations and future work.} The strategy used in the property optimization task is inapplicable for real-world drug discovery, as it is expensive to evaluate the properties of molecules with wet-lab experiments~\cite{zhu2023sample}. We believe that developing novel sample-efficient optimization approaches is a fruitful avenue. In the future, one might extend MolHF to generate other types of graphs with hierarchical structures.

\section*{Acknowledgments}

This research was partially supported by the National Key R\&D Program of China under grant No.2019YFC0118802, Key R\&D Program of Zhejiang Province under grant No.2020C03010, National Natural Science Foundation of China under grant No.62176231. Tingjun Hou’s research was supported in part by the National Natural Science Foundation of China under grant No.22220102001.

\section*{Contribution Statement}
Yiheng Zhu and Zhenqiu Ouyang contributed equally to this work.

\bibliographystyle{named}
\bibliography{ijcai23}

\clearpage
\appendix
\section{Details on MolHF}
\subsection{Criss-Cross Attention}
\label{sec:A.1}
\begin{figure}[bth]
    \centering
    \includegraphics[width=0.3\textwidth]{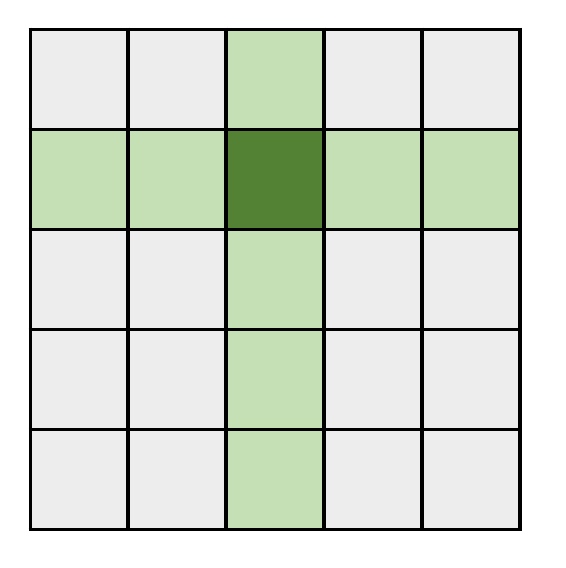}
    \caption{The visualization of contextual information in horizontal and vertical directions for a certain bond.}
    \label{fig:criss_cross_attention}
\end{figure}

We adopt Criss-Cross Attention (CCA)~\cite{huang2019ccnet} to aggregate features from the bond tensor in horizontal and vertical directions as the contextual messages of two atoms connected by a bond, respectively. For example, for the bond between the $2^{rd}$ atom and the $3^{th}$ atom, CCA can assign the $2^{rd}$ atom and the $3^{th}$ atom with features obtained from the $2^{rd}$ column bond features and the $3^{th}$ row bond features, as illustrated in Figure~\ref{fig:criss_cross_attention}.

Formally, given a bond feature maps $H \in R^{C\times N \times N}$, CCA first applies three 1 $\times$ 1 convolutional layers on $H$ to obtain queries $Q \in R^{C^{\prime} \times N \times N}$, keys $K \in R^{C^{\prime} \times N \times N}$, and values $V \in R^{C \times N \times N}$ respectively. Then we generate attention maps $A\in R^{(N + N - 1)\times N \times N}$ based on queries $Q$ and keys $K$. Finally, we use attention maps $A$ and values $V$ to calculate the contextual information of all bonds as the output $O \in R^{C \times N \times N}$ of CCA. For more comprehensive equations, we refer the readers to~\cite{huang2019ccnet}.

\section{Further Experimental Results}
\subsection{Generation and Construction}
\begin{figure}[bth]
    \centering
    \includegraphics[width=0.45\textwidth]{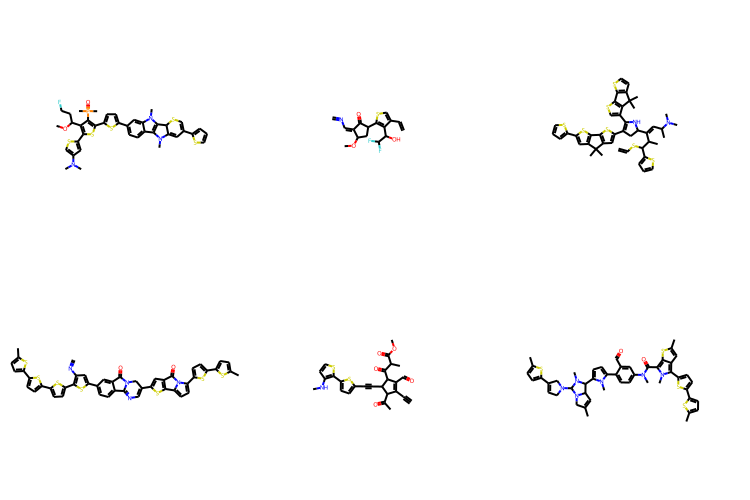}
    \caption{Illustration of molecules from MolHF on the Polymer dataset (generated according to the resampling strategy described in Section 5.3 for better sample quality).}
    \label{fig:polymer_visualization}
\end{figure}

Experiments on the Polymer dataset demonstrate that MolHF can be applied to model larger molecules that are impractical with previous flow-based models. MolHF can generate approximately $43\%$ valid molecules, most of which are large. We illustrate several valid molecules generated by MolHF in Figure~\ref{fig:polymer_visualization}.

\subsection{Hierarchical Visualization}
\begin{figure}[bth]
\centering    
\includegraphics[width=0.45\textwidth]{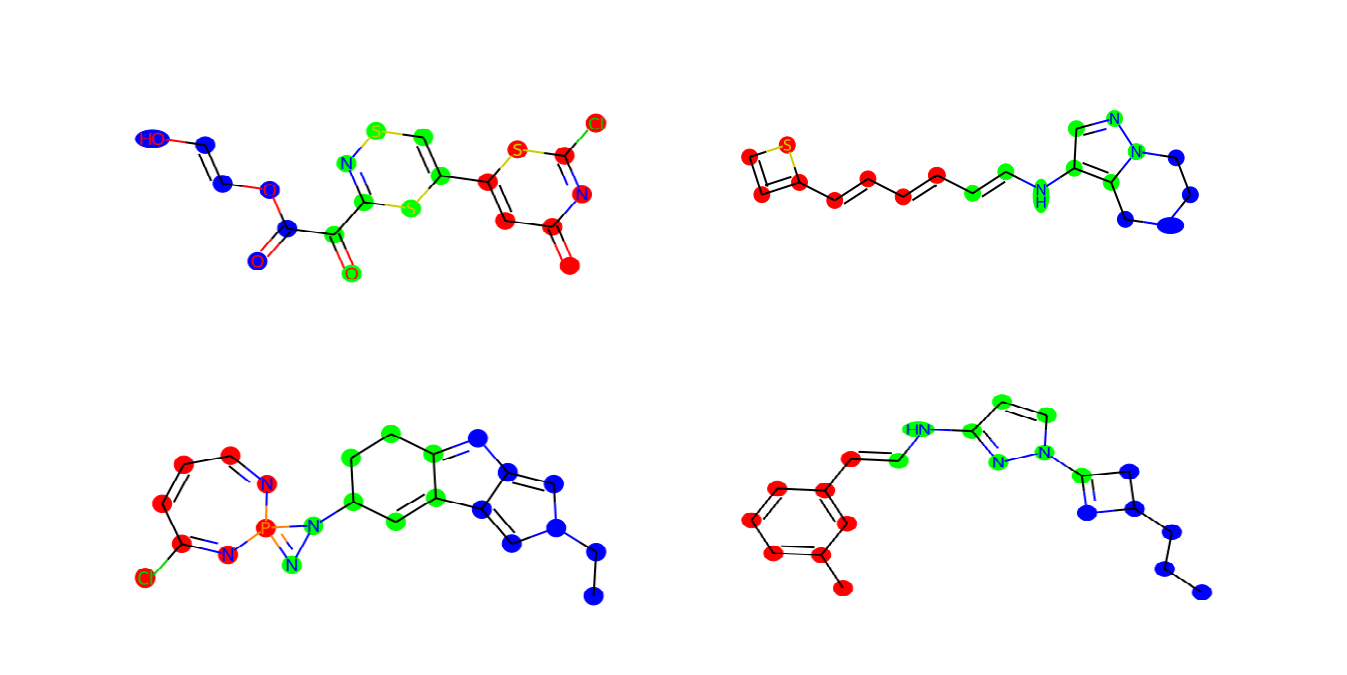} 
\caption{(a) Illustration of the most frequent substructures generated by MolHF. (b) Illustration of the generated molecules with the same coarsened graph structure, colored based on the substructures.}
\label{fig:docomposition} 
\end{figure}

To link coarsened graph structure with substructures, we show several molecules with the same coarsened graph structure in Figure~\ref{fig:docomposition}, where atoms are colored according to the substructures to which they belong. We observe that MolHF learns to hierarchically generate molecular graphs in a coarse-to-fine manner by first generating a coarsened graph structure (skeletal structure) and then refining the substructures.

\begin{table}
\centering
\begin{tabular}{lll}
\toprule
Hyper-parameters         & Range                  \\ 
\midrule
k                        & 2, 3, 4                \\
coarsened graph conditional flow step number & 6, 7, 8 \\
R-GCN layer number       & 2, 3, 4                \\
R-GCN hidden dim         & 128, 256               \\
CC-Glow flow step number & 3, 4, 5                \\
3 $\times$ 3 convolution layer hidden dim & 128, 256 \\
\bottomrule
\end{tabular}
\caption{The range of values tried per hyper-parameters.}
\label{table:range}
\end{table}

\section{Details on Implementation}

Our proposed MolHF is implemented in Pytorch (available under a BSD-style license) and trained on 1 Tesla V100 GPU. We use the RDKit (available under a BSD 3-Clause license) to process molecules. We conduct Student's t-test to analyze whether the performance difference is statistically significant. In addition, we summarize the code repositories used in this work as follows.

\begin{itemize}
    \item Glow-pytorch (https://github.com/rosinality/glow-pytorch, available under a MIT license)
    \item GCPN (https://github.com/bowenliu16/rl\_graph\_generation, available under a BSD-3-Clause license)
    \item MoFlow (https://github.com/calvin-zcx/moflow, available under a MIT license)
    \item CCNet (https://github.com/speedinghzl/CCNet, available under a MIT license)
\end{itemize}


\paragraph{Data source.} We use the following raw data sources: the ZINC250k dataset used in CVAE~\cite{gomez2018automatic} (Apache License 2.0), and the Polymer dataset used in~\cite{st2019message}.

\subsection{Hyper-parameters}
We perform a random search to tune the hyper-parameters for different datasets, the range of values tried per hyper-parameters is summarized in Table~\ref{table:range}. For the ZINC250K dataset, the coarsened graph conditional flow has 4 blocks and the CC-Glow has 3 blocks. Each block of coarsened graph conditional flow contains 6 steps where the coupling network is a 2-layer R-GCN with 256 hidden dimensions followed by a multi-layer perceptron (MLP) with 256 hidden dimensions. Each block of CC-Glow contains 3 steps where the coupling network is composed of a 3 $\times$ 3 convolution layer with 256 hidden dimensions followed by a CCA layer, followed by a 3 $\times$ 3 convolution layer. For the Polymer dataset, the coarsened graph conditional flow has 6 blocks and the CC-Glow has 5 blocks. Each block of coarsened graph conditional flow contains 8 steps where the coupling network is a 4-layer R-GCN with 128 hidden dimensions followed by a MLP with 128 hidden dimensions. Each block of CC-Glow contains 3 steps where the coupling network is composed of a 3 $\times$ 3 convolution layer with 128 hidden dimensions followed by a CCA layer, followed by a 3 $\times$ 3 convolution layer.

For molecular generation and reconstruction, we train MolHF by Adam optimizer with learning rate 0.001 and batch size 256. We train MolHF for 100 epochs on the ZINC250K and 200 epochs on the Polymer. 

For molecular property optimization, the surrogate network $h$ is a MLP with 32 hidden dimensions. We train $h$ by Adam optimizer with learning rate 0.001 and batch size 256 for 5 epochs. We set $\alpha=0.5$ for optimizing QED and PlogP.

\section{Further Discussion}
\paragraph{Negative social impact.} If intentionally misused, our proposed method may be utilized to discover environmentally harmful or toxic chemicals. Hence, it is vital to emphasize the conscious use of such generative models and propose mechanisms for monitoring misuse.

\end{document}